\documentclass{article}

\PassOptionsToPackage{numbers, compress}{natbib}

\usepackage[final]{neurips_2021}

\usepackage[utf8]{inputenc} 
\usepackage[T1]{fontenc}    
\usepackage{hyperref}       
\usepackage{url}            
\usepackage{booktabs}       
\usepackage{amsfonts}       
\usepackage{nicefrac}       
\usepackage{microtype}      
\usepackage{color}
\usepackage[table]{xcolor}

\usepackage{graphicx}
\usepackage{multirow}
\usepackage{makecell}
\usepackage{enumitem}

\usepackage{subfig}

\title{Does Thermal data make the detection systems more reliable?}

\author{%
  Shruthi Gowda, Bahram Zonooz, Elahe Arani \\
  Advanced Research Lab, NavInfo Europe, The Netherlands \\
  \texttt{\{shruthi.gowda, elahe.arani\}@navinfo.eu}, bahram.zonooz@gmail.com \\
}

\begin{document}
\maketitle

\begin{abstract}
Deep learning-based detection networks have made remarkable progress in autonomous driving systems (ADS). ADS should have reliable performance across a variety of ambient lighting and adverse weather conditions. However, luminance degradation and visual obstructions (such as glare, fog) result in poor quality images by the visual camera which leads to performance decline. To overcome these challenges, we explore the idea of leveraging a different data modality that is disparate yet complementary to the visual data. We propose a comprehensive detection system based on a multimodal-collaborative framework that learns from both RGB (from visual cameras) and thermal (from Infrared cameras) data. This framework trains two networks collaboratively and provides flexibility in learning optimal features of its own modality while also incorporating the complementary knowledge of the other. Our extensive empirical results show that while the improvement in accuracy is nominal, the value lies in challenging and extremely difficult edge cases which is crucial in safety-critical applications such as AD. We provide a holistic view of both merits and limitations of using a thermal imaging system in detection.
\footnotetext[1]{The code for this research is available at: https://github.com/NeurAI-Lab/MMC}
\end{abstract}

\section{Introduction}

Autonomous driving is a challenging and safety-critical application that has to perform reliably in an ever-changing environment. Given the ongoing discussion on sensors in ADS, one school of thought is to use visual cameras as the sole sensor. This makes consistent and reliable performance a more challenging scenario because the quality of the images captured depends a lot on the ambient lighting conditions. However, the performance of detection networks, which form a critical component in ADS, degrades with variation in ambient lighting and weather conditions.  

\begin{figure}[tb]
  \centering
  \includegraphics[width=\linewidth]{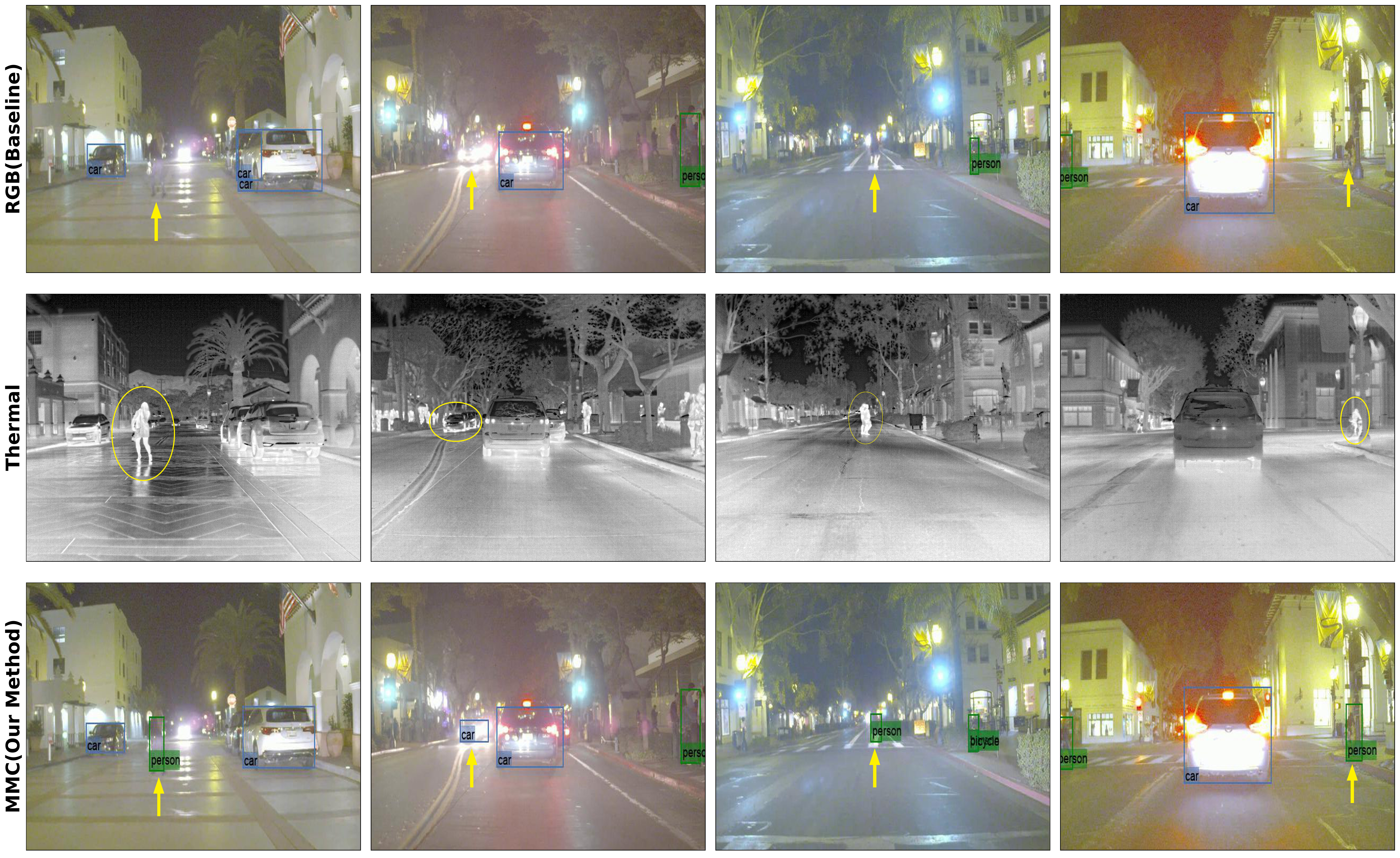}
  \caption{Examples of predictions on FLIR dataset using baseline (network trained only on RGB) and our method (MMC framework trained using both RGB and Thermal images). The addition of thermal data helps in detecting pedestrians and cars that are not clearly visible due to lighting and headlight glares as highlighted in yellow.}
  \label{fig:vis1}
\end{figure}

To improve detections in all challenging scenarios, it is favorable to leverage a data modality that is complementary to the visual RGB images. Inspired by the prior work on thermal image processing \cite{miethig2019leveraging,krivsto2020Thermal}, we explore the usage of an alternative thermal sensor that provides the information that is not captured by regular visual cameras. The Infrared (IR) cameras capture the infrared radiation emitted by objects which is dependant only on the temperature of the object, which makes it invariant to illumination, visual obstructions, and adverse weather conditions. An example is provided in the first image of Figure. \ref{fig:vis1} where the pedestrian crossing the road goes unnoticed due to the headlights in the RGB image but it is very clear in the thermal image. However, as they are less detailed and not very perceptible, thermal images alone will not suffice either. Given the nature of these modalities, RGB and thermal data are different yet complementary to each other and hence together offer more information than a single modality alone.

We explore the approach of integrating the visual data (RGB images) with the IR data (thermal images) to envision a comprehensive detection system that produces consistent detections irrespective of ambient lighting. The two modalities have different distributions but still share similar semantic information of the instances. Existing works which fuse both these modalities to learn a single representation from two considerably different distributions lead to sub-optimal solutions. We, therefore, propose a Multimodal-Collaborative (MMC) framework that collaboratively trains two networks, one for each modality, thereby allowing the flexibility in learning optimal features of its own while also incorporating the complementary knowledge of the other. We further add an auxiliary reconstruction loss to encourage the networks to exhaustively explore the input space and disentangle the texture and semantic information to learn robust representations that help in better generalization.


In summary, our contributions are as follows:
\begin{itemize}[noitemsep,nolistsep]
    \item We propose a MultiModal-Collaborative (MMC) approach for leveraging Thermal data along with RGB data for improving the generalization of detectors across varying illumination and weather conditions.
    \item We provide detailed analysis on MMC and three different techniques on two different RGB-Thermal datasets: FLIR \cite{flir} and KAIST \cite{hwang2015multispectral}.
    \item We show that MMC not only provides consistent improvement in accuracy during day and night but also improves the robustness to natural corruptions as well as targeted adversarial attacks.
    \item We provide a holistic view of both merits and limitations of the solution that can help the community in making an informed decision based on the application requirements.
\end{itemize}

\section{Related Work}

Thermal imaging domain has been explored by few prior works and  Miethig et al. \cite{miethig2019leveraging} show some of the AD fails in the past years and explore the option of leveraging thermal images to supplement the existing detection systems. Their work concentrates on introducing and comparing pedestrians, cars, and cycles in various weather and lighting conditions, in images captured by visual and IR cameras. They conclude that thermal imaging cannot be a standalone solution but can provide valuable information in certain environments when other sensors perform sub-optimally. Detection in thermal spectrum slowly gained traction in real world detection applications and Krišto et al. \cite{krivsto2020Thermal} used the Convolutional Neural Network (CNN) based object detection network to perform person detection using thermal images instead of the typical RGB images for video surveillance applications. They trained a YOLO-v3 \cite{redmon2018yolov3} detector on RGB images and then additionally also trained on thermal data but the improvements were shown by using only thermal images as test images as the application was mostly surveillance.   

Some works combined both the spectral information to help detection and the basic approach of combining the training datasets was started by Agrawal and Subramanian \cite{agrawal2019enhancing} where the network was trained on both RGB and thermal data. This simplistic approach did not provide much improvement and the addition of thermal data did not improve performance in the night-time images. A few works have performed a fusion of both spectral modalities to improve detection. Yadav et al. \cite{yadav2020cnn} proposed an architecture to fuse visual and thermal images for detection where the features from two networks are extracted and merged in the last convolution layer before feeding it to the decoder for detection. The two-stream network is computationally expensive and the simple fusion logic falls short in complex data scenarios. They show minor improvements on the KAIST \cite{hwang2015multispectral} dataset and no improvement on the FLIR \cite{flir} dataset. Multiple fusion strategies are compared in \cite{li2019illumination} where the features of RGB and thermal images are fused at different layers of the CNN network and a gate function is added at the end to weigh the contribution of the RGB and thermal sub-networks based on the illumination of the image. These methods require paired images from both modalities at inference which limits their application. Results from the above works are not shown using the standard metrics used in the detection literature which leads to difficulty in gauging the clear benefit of thermal images. 


\section{Proposed Approach}

We aim to use the information from both data modalities, as RGB images provide detailed visual cues which are complemented by the thermal images which offer missing semantic information that might be occluded or less visible in the corresponding RGB image. However, the distribution in the two modalities are quite different and therefore the optimal set of features for each would not be the same and completely sharing the features in a single network might lead to sub-optimal learned representation. Collaborative training framework, on the other hand, provides the flexibility for each network to incorporate complementary knowledge from the other modality without impeding its ability to learn the optimal representation on the modality it is trained on. 

\begin{figure}[tb]                                                                              
  \centering
  \includegraphics[width=0.9\linewidth]{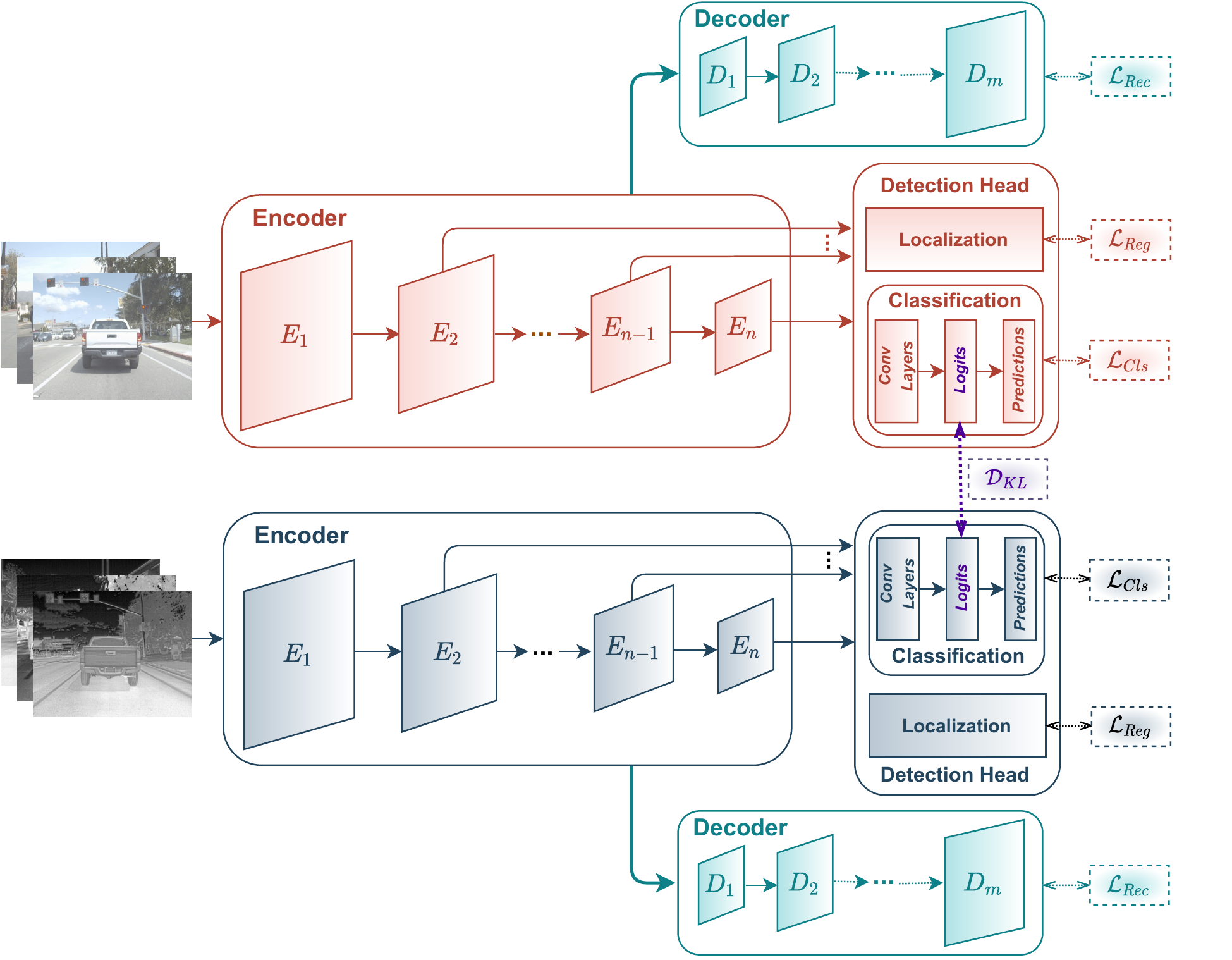}
  \caption{Schematic diagram of the MultiModal-Collaborative Learning framework. The encoder is DeiT-Tiny and SSD is the detection head. The network with red hue represents the RGB-network and the network with grey hue represents the thermal-network. The network with the green hue is the reconstruction decoder network. The networks are trained with a supervised classification ($ \mathcal{L}_{Cls}$) and regression loss ($ \mathcal{L}_{Reg}$), a mimicry loss ($D_{KL}$) and an auxiliary reconstruction loss ($ \mathcal{L}_{Rec}$).}
  \label{fig:dml}
\end{figure}

To ensure this, we employ DML \cite{zhang2018deep} which is a knowledge distillation approach that substitutes the one-way knowledge transfer from a static teacher with a cohort of student models that learn collaboratively while teaching each other. Our proposed MultiModal-Collaborative (MMC) framework uses DML as the base, with two networks each trained on a different data modality.
The RGB-network receives the RGB images while the thermal-network receives the corresponding thermal images as the input (Figure \ref{fig:dml}). Each network is trained with a supervised detection loss (standard cross-entropy and regression losses) and a mimicry loss which aligns the feature spaces of the two networks and can lead to smoother decision boundaries. We use the Kullback-Leibler divergence ($D_{KL}$) as the mimicry loss function between the networks. 

Therefore, the overall loss function per network is the sum of detection loss and mimicry loss,
\begin{equation}
   \mathcal{L}_{MMC-RGB} = \mathcal{L}_{Det} + \lambda_{rgb} \mathcal{D}_{KL}(p_{rgb} || p_{thm})
\end{equation}
\begin{equation}
   \mathcal{L}_{MMC-Thm} = \mathcal{L}_{Det} + \lambda_{thm} \mathcal{D}_{KL}(p_{thm} || p_{rgb})
\end{equation}
where $\lambda_{rgb}$ and $\lambda_{thm}$ are the balancing weights and are set to 0.1 and 1.0, respectively.

The KL divergence is applied on the soft logits $p_{rgb}$ and $p_{thm}$ which are calculated as the softmax with Temperature ($\tau=2$) introduced in \cite{hinton2015distilling}. The detection loss is a weighted summation of classification and regression losses: 
\begin{equation}
   \mathcal{L}_{Det} = \frac{1}{N_{Cls}}\mathcal{L}_{Cls} + \lambda_{Reg} \mathcal{L}_{Reg}
   \label{eqn:det}
\end{equation}
where $L_{Cls}$ is the cross-entropy loss, $L_{Reg}$ is the $l_2$ loss. $N_{Cls}$ denotes the number of classes and $\lambda_{Reg}$ is the balancing weight which is typically set to 1.


To Further encourage our method to explore the input feature space exhaustively and extract all the semantic information into the learned representations, we apply an auxiliary task for reconstructing the inputs. The auxiliary task network takes in the features from the intermediate layers of encoders and aims to reconstruct the input image via the respective decoders (Figure \ref{fig:dml}, green hue). Thus, the reconstruction losses are,
\begin{equation}
    \mathcal{L}_{Rec-RGB}= \sum (x_{rgb} - Dec_{rgb}(Enc_{rgb}(x_{rgb}))^2 
\end{equation}
\begin{equation}
    \mathcal{L}_{Rec-Thm}= \sum (x_{thm} - Dec_{thm}(Enc_{thm}(x_{thm}))^2 
\end{equation}
where $x_{rgb}$ and $x_{thm}$ are the inputs, Enc and Dec denote the Encoder and the Decoder used for feature extraction and reconstruction respectively. The total loss will be $\mathcal{L}_{MMC} + \lambda_{Rec}\mathcal{L}_{Rec}$ where the balancing weight $\lambda_{Rec}$ is 5.0 for both RGB and thermal networks.

Additionally, we perform cross-reconstruction whereby the decoder of the RGB-network reconstructs images from the features encoded by the thermal-network's encoder (Figure \ref{fig:dml_cross} in Appendix).
This encourages the backbone to disentangle texture and semantic features and learn to utilize the semantic features from a thermal image to reconstruct the corresponding RGB image. For the downstream task, the detection head selects the relevant semantic features and this helps in domain adaptation as the semantic features remain the same during different lighting conditions. 

The cross-reconstruction losses are given by,
\begin{equation}
    \mathcal{L}_{CrossRec-RGB}= \sum (x_{rgb} - Dec_{rgb}(Enc_{thm}(x_{thm}))^2 
\end{equation}
\begin{equation}
    \mathcal{L}_{CrossRec-Thm}= \sum (x_{thm} - Dec_{thm}(Enc_{rgb}(x_{rgb}))^2 
\end{equation}
The total loss is $\mathcal{L}_{MMC} + \lambda_{CrossRec}\mathcal{L}_{CrossRec}$, where the balancing weight $\lambda_{CrossRec}$ is 10.0 and 5.0 for RGB and thermal networks, respectively.

\section{Experiment}
\label{baseline}

\subsection{Object Detection Architecture}
The detection network consists of an encoder to extract features and a detection head for classification and bounding-box regression. Transformers have paved their way into computer vision applications with networks such as Vision Transformer(ViT) \cite{dosovitskiy2020image} and Data-efficient Transformers (DeiT) \cite{deit}. Ze Liu et al. \cite{liu2021swin} show improved performance in detection with transformer plugged in as the backbone. We use DeiT-Tiny as the encoder network. The architecture consists of repeated blocks of self-attention, feedforward layers, and an additional distillation module. To extract meaningful image representations, we extract the learned embeddings from the final Transformer block and add an extra block to get features at different scales before sending it to the detection head. Single Shot Multibox Detector (SSD) \cite{ssd} is used as the detection head and performs localization and classification in a single forward pass and uses predefined anchors to make predictions. SSD employs a Feature Pyramid Network (FPN) and uses six feature maps with progressively reducing resolutions to facilitate detecting objects at multiple scales and overall provides a better trade-off between speed and accuracy.

\subsection{Baseline}
We compare our method with three baselines: (1) RGB images only, (2) Thermal images only, and (3) RGB + Thermal (both RGB and thermal data combined into a new dataset) to train a single network.

\subsection{Fusion}  
We also compare our method with two different fusion techniques to integrate visual and thermal modalities: Input Fusion and Feature Fusion (Figure \ref{fig:fusion} in Appendix). In the Input Fusion, the visual and thermal image pairs are concatenated before feeding them to the network. The first convolution block needs to be changed to accommodate the increase in the input channel size. On the other hand, Feature Fusion integrates the features at higher layers of encoders, which are semantically richer. 
In the Feature Fusion, there are two transformer encoder networks and one detector head network. One of the encoders receives color images and the other receives the corresponding thermal images as the input. The features at multiple levels are concatenated (channel-wise) followed by a Network-in-Network module \cite{lin2013network} for dimension reduction and the concatenated features are then fed to the detection head.

\subsection{Style-Transfer}
Inspired by the recent ideas of style transfer from one type of image to another \cite{huang2017arbitrary}, we explore the idea of transferring the style of thermal modality onto the visual modality. We use the Adaptive Instance Normalization (AdaIN) method for style transfer. Given a content image (RGB) and style image (thermal), ADaIN combines the content of the former with the style of the latter by transferring certain feature statistics (mean and variance). This can be achieved even with a small subset of thermal images as style images. We transfer the 'thermal-style' onto the training RGB images and create a new training set, thermal-stylized-RGB (see Figure \ref{fig:style} in Appendix for example images). We train a single network by combining both the original RGB and the thermal-stylized-RGB images.

\section{Emperical Validation}
We briefly introduce the datasets, the evaluation metrics, and the experimental setup used for the experiments.

\subsection{Datasets}
KAIST \cite{hwang2015multispectral} is a multi-spectral dataset that provides aligned RGB-Thermal images captured in day/night traffic scenes in Korea. The categories annotated in this dataset are person, people, and cyclist and are mainly used for pedestrian detection. There are three sets each for the daytime and night captured on campus, road, and downtown. So there are six sets in total for both training and testing data. According to Li et al. \cite{limultispectral}, the original annotations included few errors and were also redundant, and hence, they proposed a sanitized version of the training annotations. We use these sanitized set for our experiments which contain 7601 visual-Thermal pairs for training and 2252 pairs for testing. 

FLIR \cite{flir} is a multi-spectral dataset created by the company FLIR. The dataset has 10228 frames and 9214 annotations of five different categories: person, car, bicycle, dog, and other vehicle. The dataset comprises 60\% day and 40\% night images captured while driving in California. The annotations are only available for thermal images and not their visual counterparts. The FLIR dataset, although available in pairs, is not completely aligned as both the cameras had a different field of view while capturing the same scene. Hence, the same annotations could not be used for RGB images. We used a pre-trained detection model to infer on the RGB images and created the annotations ourselves and used these balanced set of annotations for our experiments. These annotations can be made available upon request.

\subsection{Evaluation Metrics}
We use the mean Average Precision (mAP) and F1 score as the accuracy metrics as these are the metrics in all state-of-the-art detection networks. mAP requires a series of precision-recall curves with the IoU threshold set at varying levels of difficulty. 
Also, given the precision and recall values at the segmented intervals, the F1 score is computed at recall value $0.5$ as follows:
\begin{equation}
    F1_{0.5} = 2 \times \left(\frac{precision_{0.5} \times recall_{0.5}}{precision_{0.5} + recall_{0.5}} \right)
    \label{eqn:f1}
\end{equation}

\subsection{Experimental Setup}
\label{exp}
The complete framework is implemented in Pytorch 1.7  \cite{pytorch}. Default PyTorch weight initialization, with a fixed seed value, is used for the detection head and pre-trained ImageNet weights are used for the encoder network. For data augmentation, we use random crop, and random photometric distortions which include random contrast within the range of [0.5, 1.5], saturation [0.5, 1.5], and hue [-18, +18]. 

We use a batch size of 16 and train the models with AdamW optimizer \cite{AdamW} with an initial learning rate of 5e-4 with 0.5 weight decay. For all the experiments, we evaluate the models on an NVIDIA RTX 2080Ti GPU. 


\section{Result}

Table \ref{tbl:flir_kaist} shows the results of our proposed MMC methods for KAIST and FLIR datasets along with the results of baselines, fusion, and stylization models (explained in Section \ref{baseline}) for comparison. The test set for all the experiments consists of two sets: RGB and Thermal. The RGB test set is further divided into Day and Night images, to analyze the effect of thermal images during different times of the day. The focus is on RGB test data as these images are the ones used in the end application. The inference is performed on one network (RGB-network in Figure \ref{fig:dml}) for all the experiments. (Results on the thermal test set are also included for completeness but network trained only on thermal data performs well on it.)

\begin{table}[tb]
\centering
\resizebox{\textwidth}{!}{%
\begin{tabular}{|l|l|cc|cc|cc|cc|}
\hline
\multirow{3}{*}{\rotatebox[origin=c]{90}{Dataset}} & \multirow{3}{*}{\textbf{Method}} & \multicolumn{6}{c|}{\textbf{Test\_RGB}} & \multicolumn{2}{c|}{\textbf{Test\_Thermal}} \\ \cline{3-10} 
 &  & \multicolumn{2}{c|}{Test\_All} & \multicolumn{2}{c|}{Test\_Day} & \multicolumn{2}{c|}{Test\_Night} & \multicolumn{2}{c|}{Test\_All} \\ \cline{3-10} 
 &  & mAP & F1 Score & mAP & F1 Score & mAP & F1 Score & mAP & F1 Score \\ \hline
\multirow{9}{*}{\rotatebox[origin=c]{90}{\textbf{KAIST}}} & RGB & 9.59 & 14.95 & 11.91 & 17.81 & 5.52 & 9.39 & 0.70 & 1.34 \\
 & Thermal & 0.49 & 1.00 & 0.63 & 1.22 & 0.30 & 0.60 & \textbf{16.52} & \textbf{24.01} \\
 & RGB + Thermal & 9.11 & 14.29 & 11.20 & 17.20 & 5.91 & 9.66 & 6.81 & 11.48 \\ \cline{2-10} 
 & Input Fusion & 3.25 & 5.79 & 4.40 & 7.68 & 1.38 & 2.60 & 0.12 & 0.23 \\
 & Feature Fusion & 9.06 & 14.15 & 11.80 & 17.77 & 4.89 & 8.24 & 0.63 & 1.21 \\ \cline{2-10} 
 & \begin{tabular}[c]{@{}l@{}}RGB + \\ Thermal stylized RGB\end{tabular} & 8.24 & 13.08 & 10.15 & 15.61 & 4.90 & 8.34 & 0.25 & 0.50 \\ \cline{2-10} 
 & MMC & 9.85 & 15.20 & 12.50 & 18.42 & \textbf{6.05} & \textbf{10.10} & 0.48 & 1.00 \\
 & MMC + Recon & \textbf{10.46} & \textbf{16.00} & \textbf{12.98} & \textbf{19.25} & 6.00 & 9.95 & 0.82 & 1.57 \\
 & MMC + Cross Recon & 10.19 & 15.5 & 12.35 & 18.31 & 5.99 & 9.65 & 0.61 & 1.20 \\ \hline \hline
\multirow{9}{*}{\rotatebox[origin=c]{90}{\textbf{FLIR}}} & RGB & 69.87 & 80.19 & 70.97 & 80.92 & 67.77 & 78.46 & 20.57 & 27.41 \\
 & Thermal & 31.48 & 43.34 & 36.70 & 49.08 & 18.29 & 27.13 & \textbf{40.97} & \textbf{50.24} \\
 & RGB + Thermal & 69.40 & 79.42 & 69.98 & 79.82 & 68.64 & 78.40 & 6.03 & 8.82 \\ \cline{2-10} 
 & Input Fusion & 29.01 & 41.68 & 32.13 & 45.17 & 20.84 & 31.04 & 0.01 & 0.01 \\
 & Feature Fusion & 41.21 & 54.46 & 44.96 & 58.15 & 28.04 & 39.65 & 0.01 & 0.01 \\ \cline{2-10} 
 & \begin{tabular}[c]{@{}l@{}}RGB + \\ Thermal stylized RGB\end{tabular} & 67.91 & 78.5 & 69.35 & 79.52 & 65.30 & 75.99 & 0.85 & 1.67 \\ \cline{2-10} 
 & MMC & 70.01 & 80.23 & 71.20 & \textbf{81.57} & \textbf{69.62} & \textbf{79.03} & 0.59 & 1.17 \\
 & MMC + Recon & \textbf{70.73} & \textbf{80.62} & \textbf{71.86} & 81.37 & 67.91 & 78.17 & 0.03 & 0.06 \\
 & MMC + Cross Recon & 65.72 & 76.59 & 67.18 & 77.65 & 63.25 & 73.93 & 0.09 & 0.17 \\ \hline
\end{tabular}
}
\caption{Accuracy results on KAIST and FLIR dataset using mAP@0.5 IoU and the F1 Score, respectively. The first three groups represent the baselines, Fusion methods and Stylization Methods which are trained on single networks (except Feature Fusion). The last grouping is our MMC approach with two networks. MMC methods show improvements on day, night and the overall RGB test set and the highest results are shown in bold.}
\label{tbl:flir_kaist}
\end{table}

The RGB+Thermal baseline shows that training with a combination of RGB and thermal data does improve accuracy over just using RGB images on nighttime images. The Input Fusion method does not perform well as fusing raw pixel data is not useful in extracting and combining any relevant attributes from the different spectral domains. Feature Fusion combines the semantic information and does better than Input Fusion but still is lower than baseline because forcing a network to learn single representation for different distributions is not an optimal solution. The fusion techniques do particularly worse on FLIR dataset because the image pairs between RGB and thermal are not registered which causes a mismatch while merging the corresponding image pixels or features. The style-transfer serves as an augmentation step that transfers the style of a thermal image into the RGB images but the results do not improve over baseline.

MMC techniques collaboratively train two networks and are the only methods to show improvement on the day, night, and the overall test set for both datasets. MMC outperforms on nighttime images and MMC with reconstruction achieves the best result for day-time images. For the night scenes, the mAP increases by a margin of ~1 mAP on KAIST and ~2 mAP on FLIR dataset. The cross-reconstruction fails to perform well on FLIR dataset (\ref{tbl:flir_kaist}) because the images are not registered and hence the features extracted from one modality fail to produce an accurate reconstruction in the other.

\subsection{Robustness to Corruption}

ADS needs to be robust to ever-changing environments and function well, come rain or shine. To test the performance of detection networks on different challenging scenarios such as varying weather conditions, sensors, and other external influences, we create a dataset by adding natural corruptions to the RGB test set. Following \cite{michaelis2019benchmarking}, we use fifteen different corruptions categorized into four groups: Noise, Blur, Weather, and Digital effects. Noise consists of Gaussian, Impulse, and Shot noise. The Blur group consists of Defocus, Glass, Motion, and Zoom blur effects. We use Brightness, Fog, Frost, and Snow to mimic different weather conditions. Finally, we account for Digital effects by adding changes in Contrast, Elastic Transformation, JPEG compression, and Pixelation. These 15 corruptions are applied at severity level 3, with level 1 being less severe and 5 being most severe corruption. (Figure \ref{fig:corrupt_visual} in Appendix)


Figure \ref{fig:kaist_corrupt} shows the accuracy results for 15 different corruptions on KAIST and FLIR datasets, respectively. The MMC methods perform better than baselines on almost all the corruptions for both datasets. On different weather conditions, e.g, snow, we get $\sim$3 mAP improvement on FLIR and $\sim$1 mAP improvement on the rest. The MMC cross reconstruction method outperforms the rest and in the noisy scenes, FLIR shows $\sim$13 mAP improvement in images with gaussian noise. We see $\sim$2 mAP on blurry images of KAIST and the improvement is marginal on the digital effects. The performance gain can be attributed to smoother decision boundaries and auxiliary reconstruction tasks in the collaborative framework which helps in better generalization.


\begin{figure}[tb] 
    \centering
    \begin{tabular}{cc}
         \includegraphics[width=\textwidth]{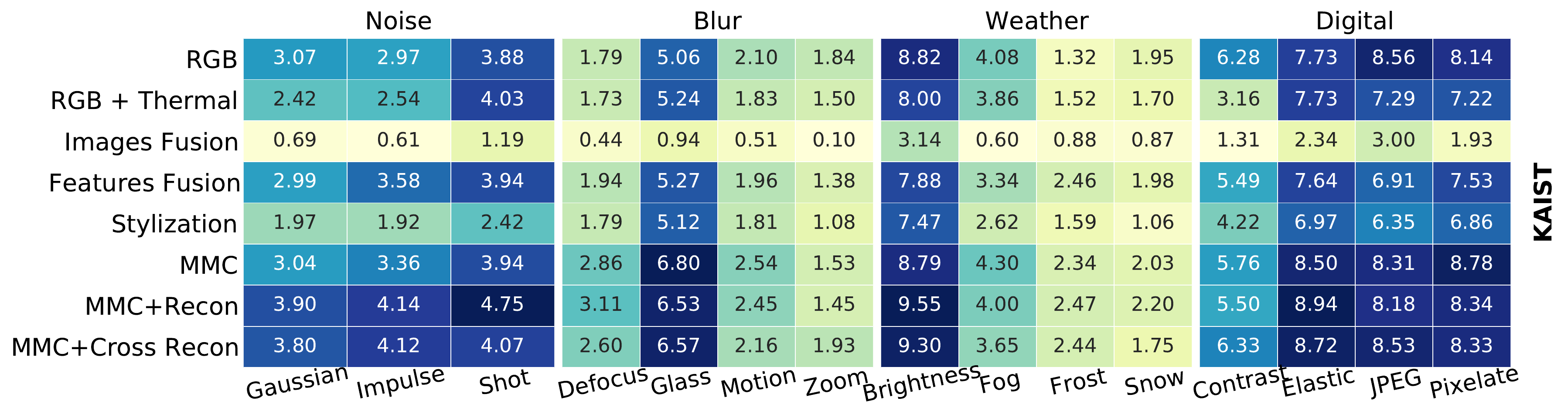} \\ \includegraphics[width=\textwidth]{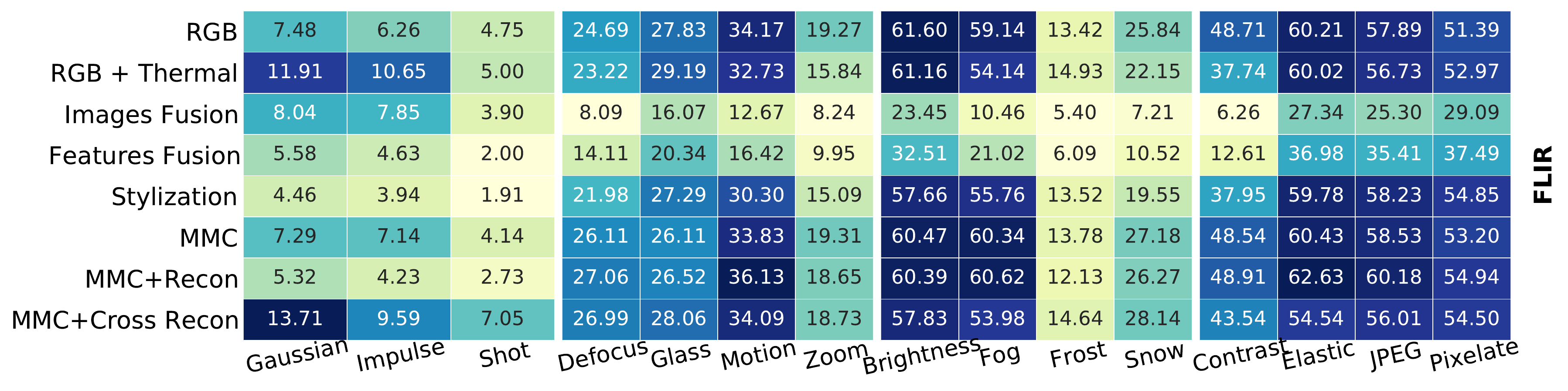} \\
    \end{tabular}
    \caption{Accuracy (mAP) results of all the methods on corrupted KAIST and FLIR datasets for fifteen different natural corruptions. MMC methods show improvement across all the categories.}
    \label{fig:kaist_corrupt}%
\end{figure}

\subsection{Robustness to Adversarial Attacks}

\begin{figure}[tb] 
    \centering
    \begin{tabular}{cc}
         \includegraphics[width=0.4\textwidth]{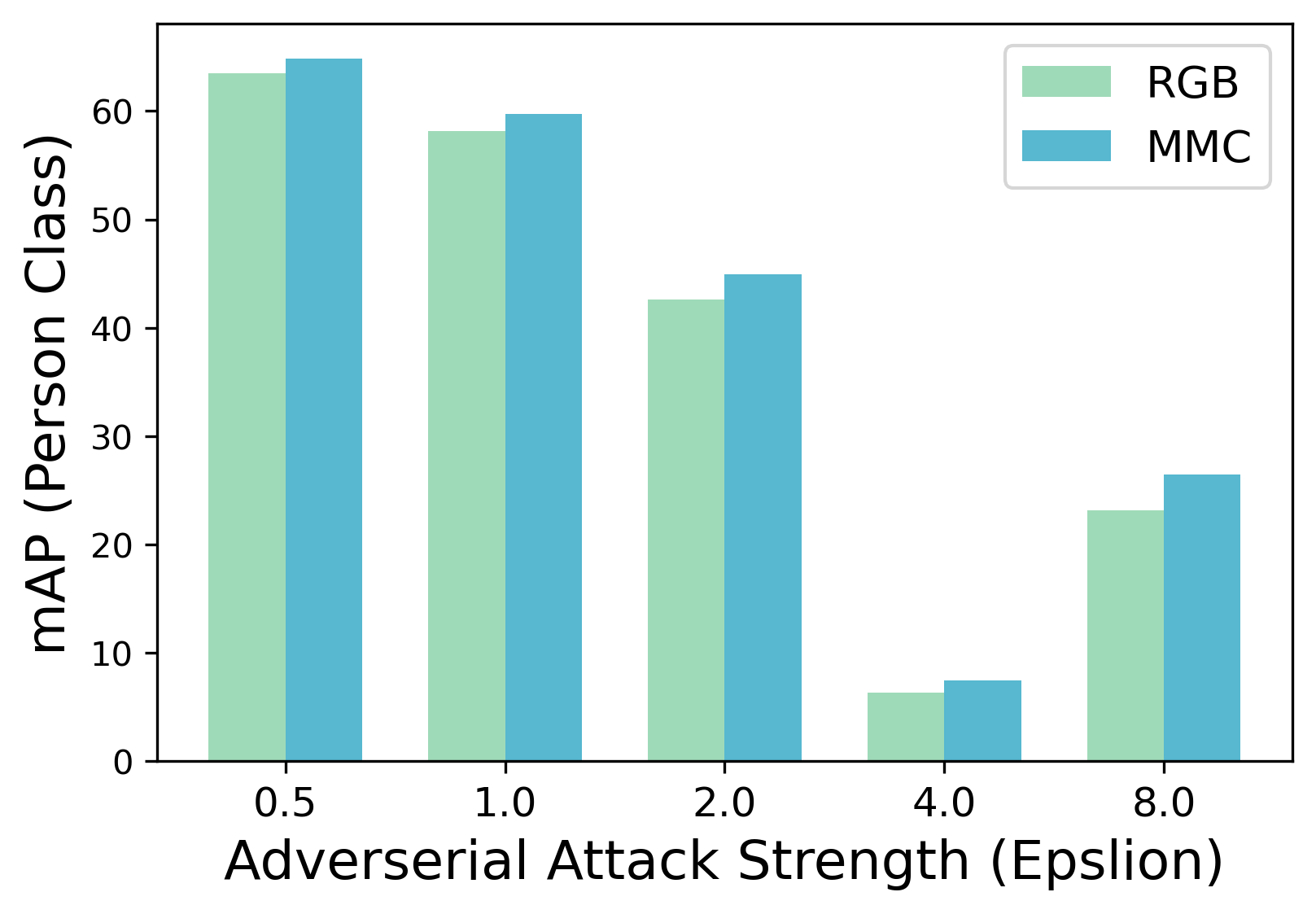} & \includegraphics[width=0.4\textwidth]{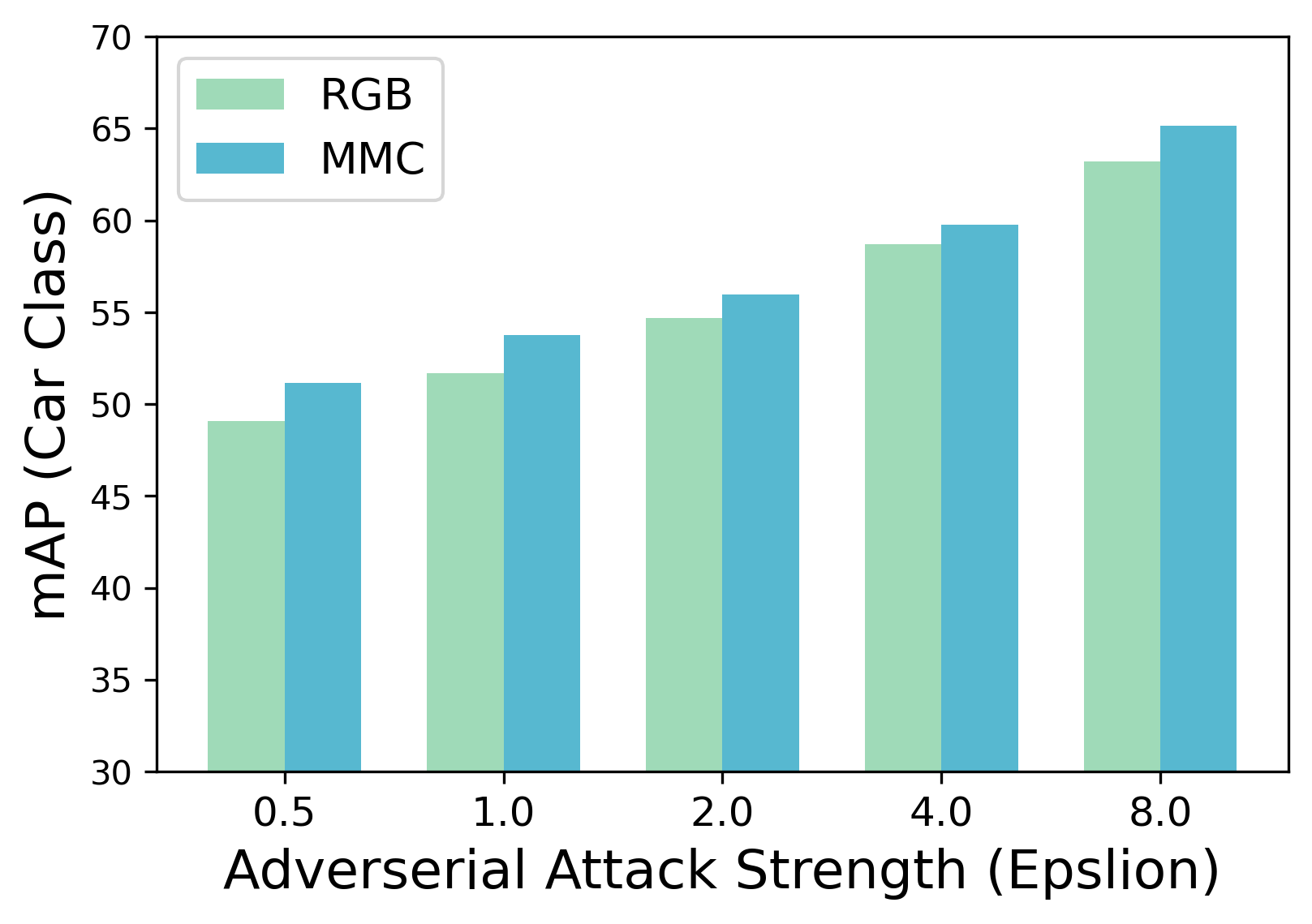} \\
         \footnotesize{(a) Targeted attack by hiding the class "Person"} & \footnotesize{(b) Targeted attack by hiding the class "Car"}
    \end{tabular}
    \caption{Robustness to targeted attacks on the FLIR dataset. MMC consistently shows improvement across varying degrees of the attack.}
    \label{fig:atack}%
\end{figure}

Deep Neural networks are shown to be vulnerable to adversarial attacks which are carefully crafted imperceptible noise added to the input image to fool the network. These can have disastrous consequences for security-critical applications like ADS which can be fooled to consider a stop sign as 100 km/h speed limit and make untimely decisions. Amongst the adversarial attacks, gradient-based attacks have access to the network gradients and use them to generate a perturbation vector. We consider the Projected Gradient Descent (PGD) \cite{mkadry2017towards} which is a first-order adversary utilizing the local first-order information about the network to generate an adversarial perturbation within an epsilon bound. To analyze how robust thermal data is against adversarial cases, we consider a plausible case scenario where pedestrians/vehicles are less visible or camouflaged by their background and design a targeted attack by hiding certain classes. To this end, we generate a target image with these modifications and use the PGD attack to generate adversarial perturbation which minimizes the loss of the model on these modified targets (Figure \ref{fig:attack_visual}). For the first experiment, we hide the class "Person" by changing the Person label to the background and for the second experiment, we hide the class "Car". Figure \ref{fig:atack} shows the results of both experiments for varying epsilon values on FLIR dataset (KAIST was not considered for this study as it does not have the "car" class). In both cases, the MMC approach performs marginally better and the improvement stays consistent through various degrees of the attack. Collaborative learning enables for smoother decision boundaries which result in improved robustness of the networks \cite{zhang2019theoretically}. 



\section{Discussion and Conclusion}
\label{conclusion}
We addressed the shortcoming of detection networks to perform reliably across different domains such as day, night, and varying weather conditions, which is a safety concern in AD applications. To help solve this domain gap, we explored the idea of leveraging data from the thermal sensor to fill in the missing details of the regular visual cameras. To this end, we proposed a MultiModal-collaborative (MMC) framework that learns between two data modalities in a collaborative manner and also extend the framework to include auxiliary reconstruction tasks. We also reported results from other different approaches of combining RGB and thermal images for comparison. MMC-based methods show consistent improvement over the test data across both datasets and also show higher robustness against corruptions and adversarial attacks. 


\begin{figure}[tb] 
    \centering
    \begin{tabular}{cc}
         \includegraphics[width=0.4\textwidth]{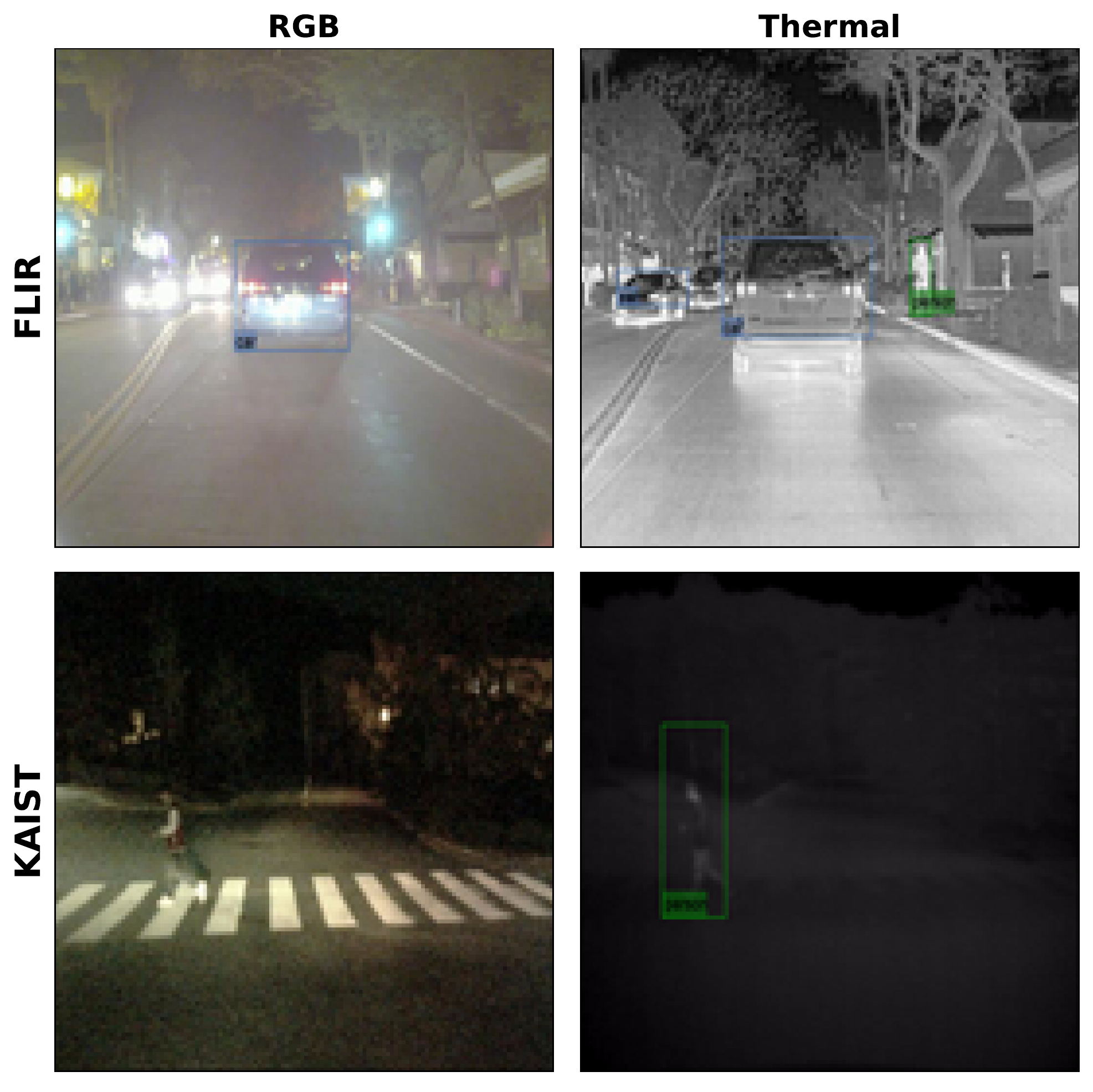} & \includegraphics[width=0.4\textwidth]{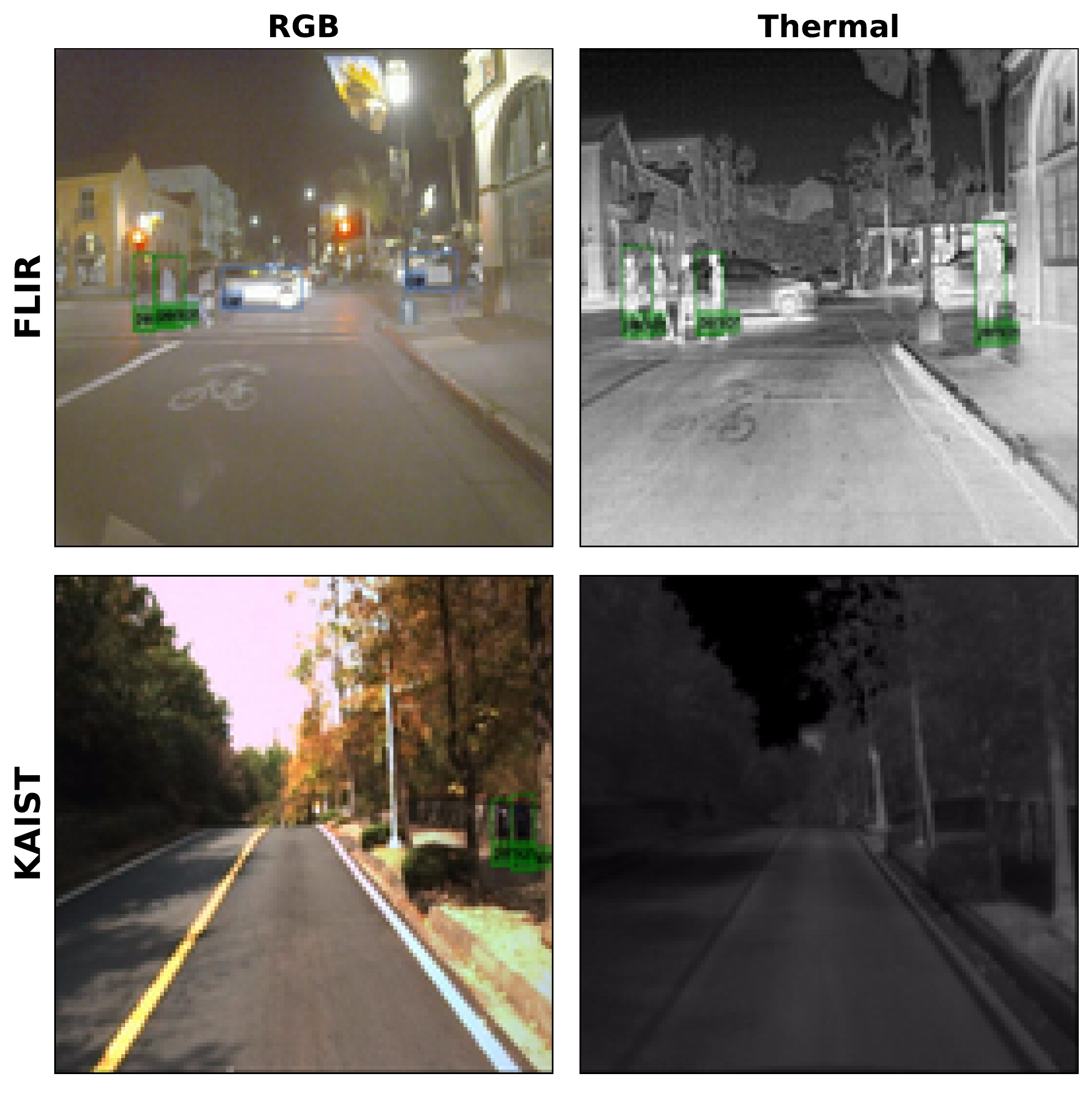} \\
        \footnotesize{(a) Examples of instances only predicted in thermal data}  & \footnotesize{(b) Examples of instances missed in thermal data}
    \end{tabular}
    \caption{Predictions on RGB-thermal image pairs for FLIR and KAIST Dataset. (a) samples when thermal images aids in detection (b) samples when RGB images alone suffices.
    Note: FLIR image pairs are not perfectly registered as shown in the images.}
    \label{fig:limit}%
\end{figure}

For a complete evaluation of the solution proposed in this work, we also address some overheads of using IR cameras. The IR cameras available are more expensive compared to their visual counterparts and hence, cost-effectiveness is an important criterion of consideration. The RGB-thermal image pairs also need to be perfectly registered which adds additional overhead. We were limited to report results on only two datasets because these are the only two available publicly. Amongst them, FLIR is not perfectly aligned and the available annotations of thermal data could not be used for RGB. We also highlight the edge cases when thermal images add value alongside a few scenarios where it offers no extra information. In Figure \ref{fig:limit}(a), a model trained on thermal data can detect the two cars while the model trained on RGB data misses it as it is obstructed due to the headlights on FLIR dataset. Thermal also helps in detecting the camouflaging pedestrian in the night while RGB model fails on KAIST dataset. But, in Figure \ref{fig:limit}(b), there are some easy detections that the thermal model fails to identify while the RGB model does better. Hence, in applications where safety is not the utmost criteria, thermal might not add more value to the existing visual data and might cause extra cost and computational overhead. On the other hand, in ADS where even a single extra detection can help avoid a disastrous accident, thermal data proves to be more beneficial. We hope to have provided a complete picture of the improvements along with the limitations which can help the AI community in considering this solution based on the requirements of the application. 


\section{Broader Impact}
\label{impact}
In this work, we provide detailed insights into the idea of using a different sensor than the regular visual cameras. We also provide a holistic view of the merits and limitations of this dual imaging system. Our extensive evaluation and findings can act as a guideline for the industrial community planning to invest in a different camera sensor as a solution for scene understanding to gauge the trade-offs between performance and reliability improvements and the additional overheads in order to make an informed choice for a particular application. Based on the findings, this solution can be applied to any safety-critical applications which face inconsistencies owing to low-quality images such as surveillance and ADS. We hope our work plays a part in the ultimate goal of creating safer and more robust AI systems.


\bibliographystyle{ieeetr}
\bibliography{references}

\newpage
\appendix
\section{Appendix}



\begin{figure}[!ht]                                                                              
  \centering
  \includegraphics[width=0.9\linewidth]{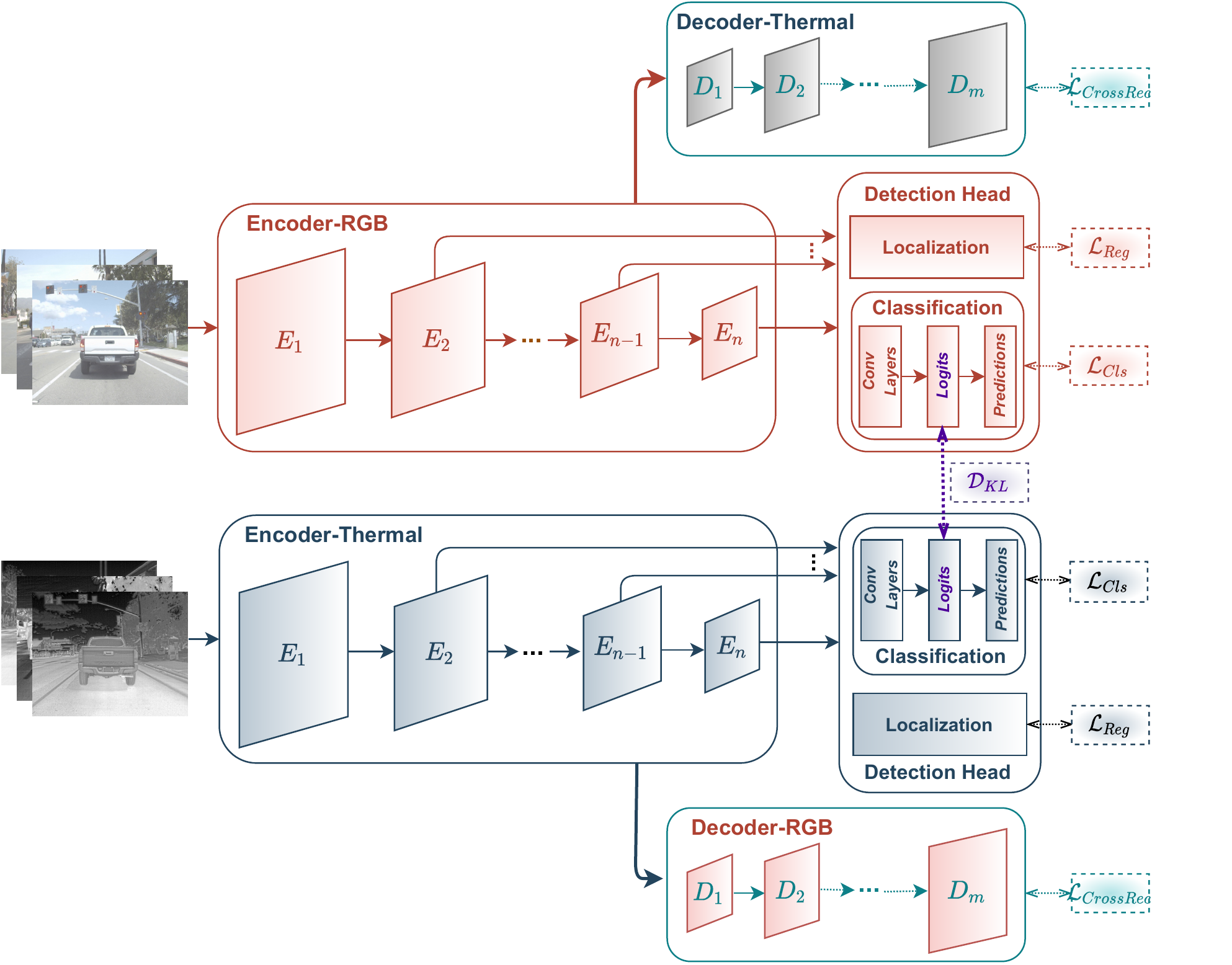}
  \caption{Schematic diagram of the MultiModal-Collaborative Learning framework with Cross Reconstruction. The encoder is DeiT-Tiny and SSD is the detection head. In MMC+Cross Reconstruction variant, the intermediate features of the RGB image (i.e, features from RGB encoder) are passed to the thermal decoder to reconstruct the original RGB image and vice-versa. The networks are trained with a supervised classification ($ \mathcal{L}_{Cls}$) and regression loss ($ \mathcal{L}_{Reg}$), a mimicry loss ($D_{KL}$) and an auxiliary cross reconstruction loss ($ \mathcal{L}_{CrossRec}$).}
  \label{fig:dml_cross}
\end{figure}

\begin{figure}[!ht]
  \centering
  \includegraphics[width=\linewidth]{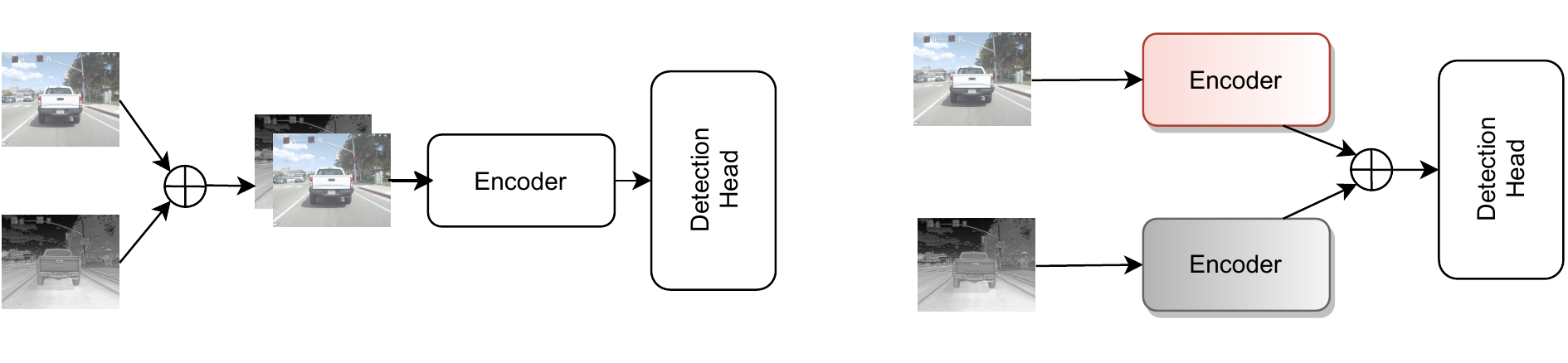}
  \caption{Schematic diagram of the Input Fusion and Feature Fusion methods respectively. (a) Input Fusion - combines RGB and thermal into a new dataset and trains the network (b) Feature Fusion - concatenates features from RGB and Thermal Encoders and trains the detection head}
  \label{fig:fusion}
\end{figure}


\begin{figure*}[!ht] 
    \centering
    \includegraphics[width=0.9\textwidth]{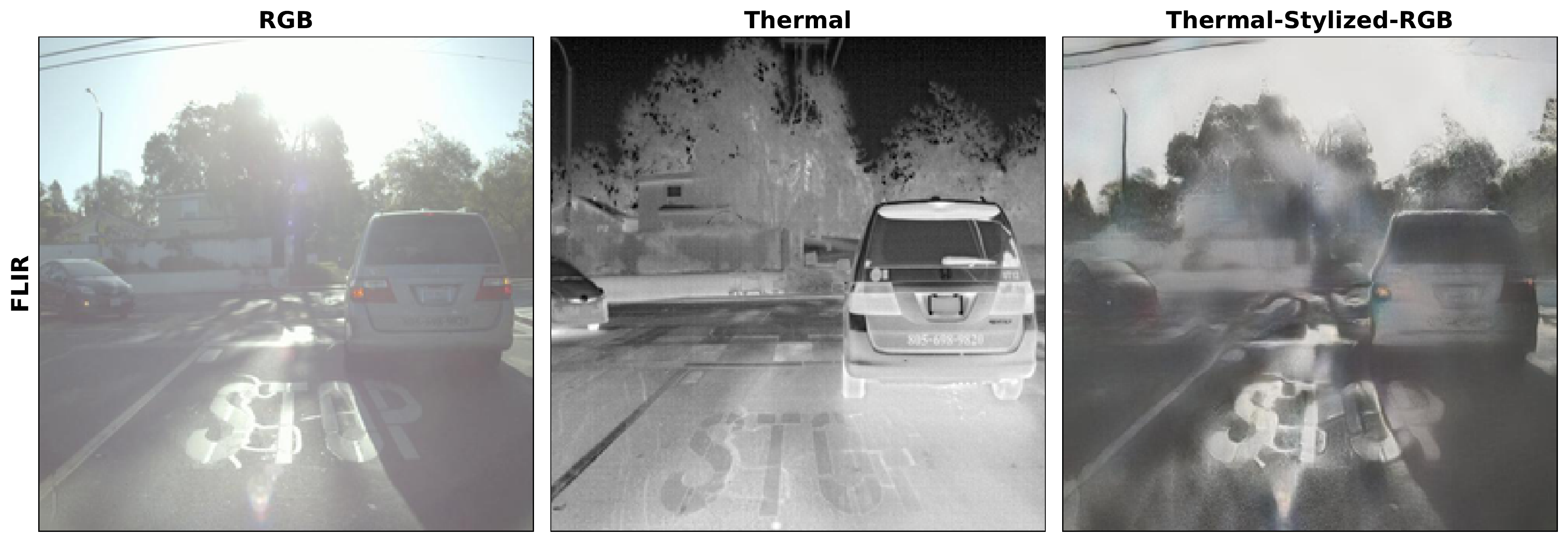} \\ \includegraphics[width=0.9\textwidth]{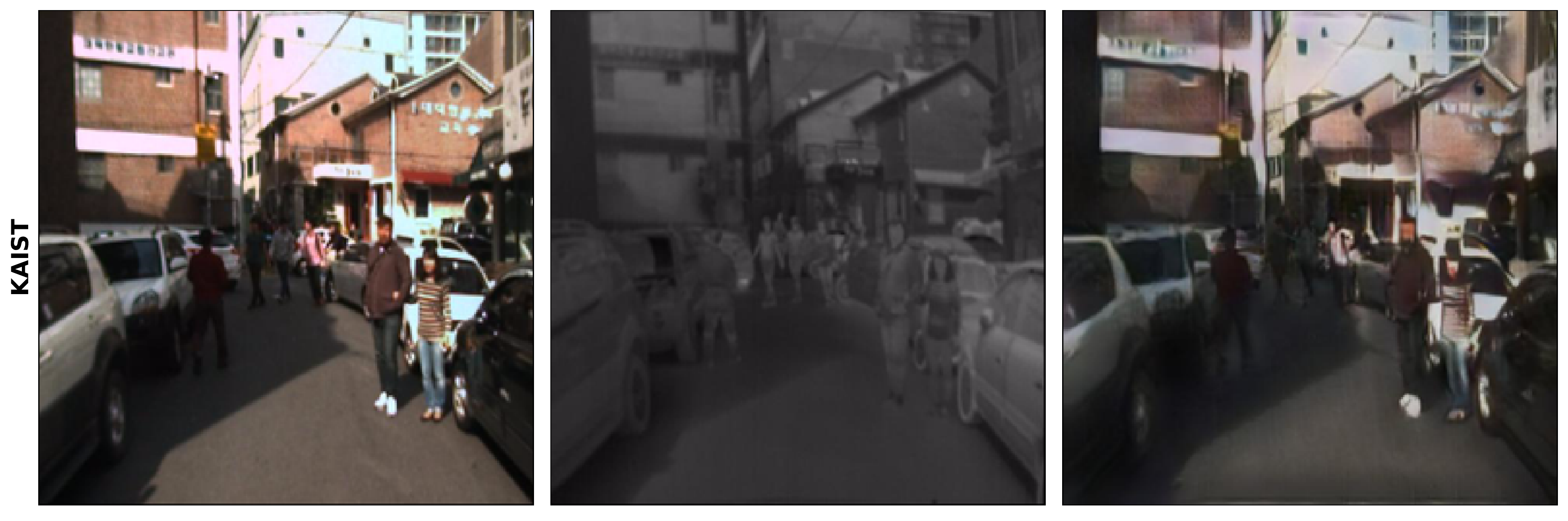} \\
    \caption{Examples of transferring style from Thermal images onto RGB images}
    \label{fig:style}%
\end{figure*}





\begin{figure}[!ht]
  \centering
  \includegraphics[width=\linewidth]{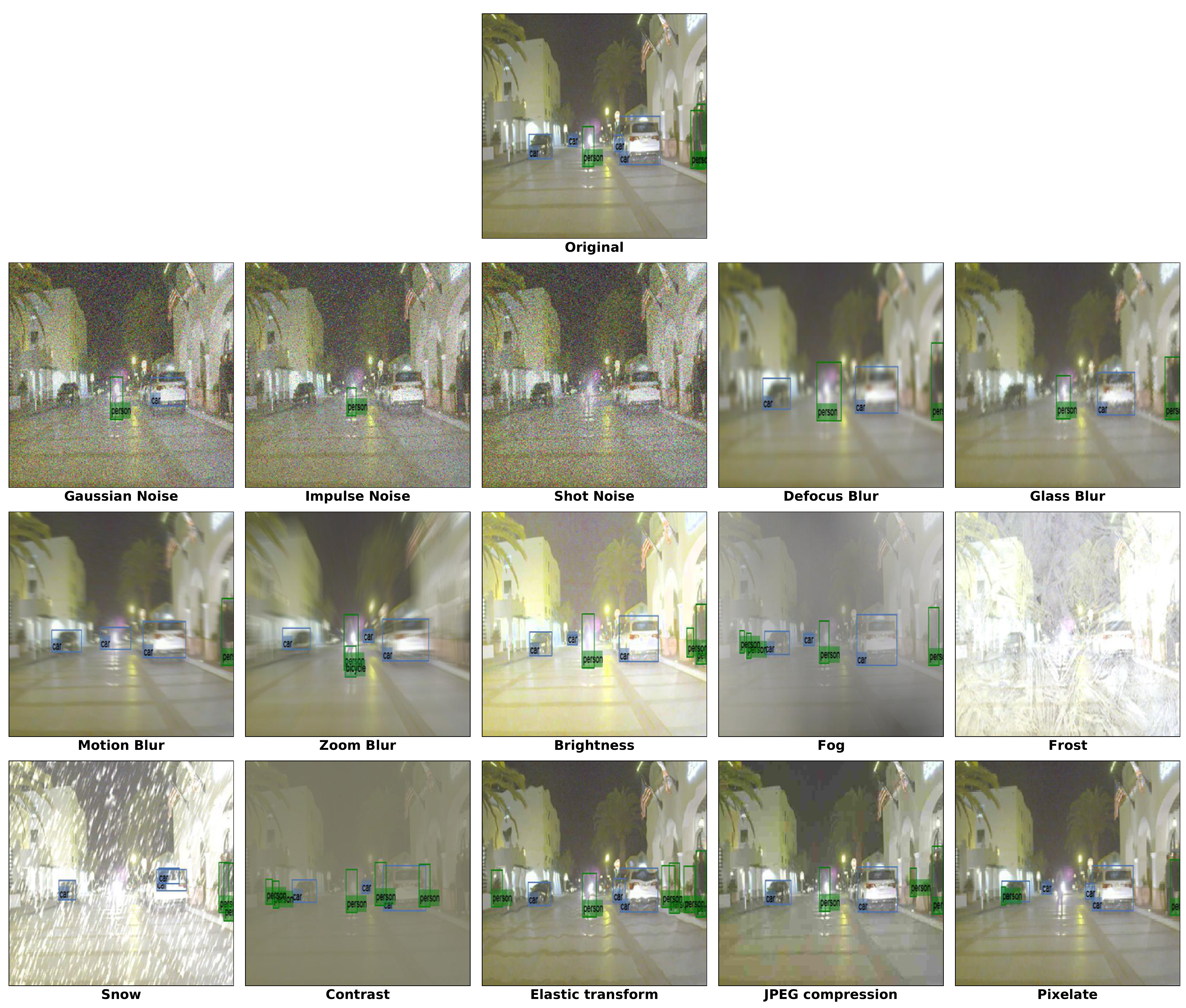}
  \caption{Predictions of MMC + Cross Reconstruction approach on the corrupted FLIR dataset. }
  \label{fig:corrupt_visual}
\end{figure}

\begin{figure}[tb]
  \centering
  \includegraphics[width=\linewidth]{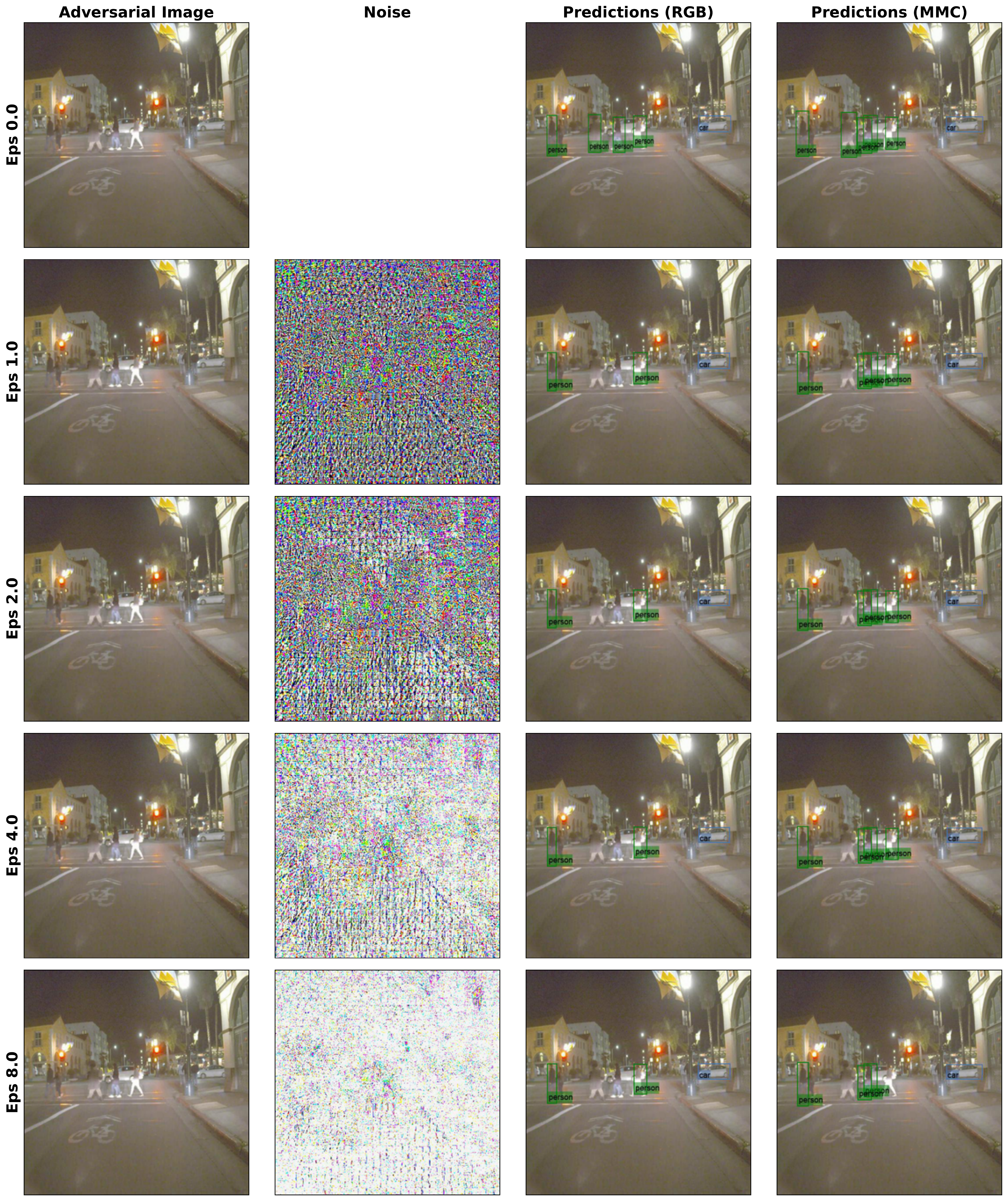}
  \caption{Predictions on FLIR images attacked by hiding the class "Person". The first column show the adverserial images created by adding the noise in the second column to the input images. Third column shows the prediction of baseline network on the adversarial images and the last column shows the predictions of the proposed MMC method. MMC shows better consistent predictions of class 'person' even with increase in attack strength}
  \label{fig:attack_visual}
\end{figure}


\end{document}